\documentclass[runningheads]{llncs}

\usepackage{microtype}
\usepackage{graphicx}
\usepackage{booktabs} 
\usepackage{hyperref}

\usepackage{float}
\usepackage{url}            
\usepackage{booktabs}       
\usepackage{amsfonts}       
\usepackage{nicefrac}       
\usepackage{microtype}      
\usepackage{comment}
\usepackage{amsmath}
\usepackage{algorithm}
\usepackage{algorithmic}
\usepackage{graphicx}
\usepackage{enumitem}

\usepackage{subcaption}
\usepackage[sort]{cite}
\usepackage{wrapfig}

\def\etal{{\em et al.\;}}

\begin{document}
\title{Policy Prediction Network: \\
Model-Free Behavior Policy with Model-Based
Learning in Continuous Action Space}
\titlerunning{Policy Prediction Network}
\author{Zac Wellmer \inst{1} \and James T. Kwok \inst{1}}
\institute{Hong Kong University of Science and Technology, Clear Water Bay, Hong Kong \\
\email{zac@1984.ai}}
\maketitle

\begin{abstract}
This paper proposes a novel deep reinforcement learning architecture that was inspired by previous tree structured architectures which were only useable in discrete action spaces. Policy Prediction Network offers a way to improve sample complexity and performance on continuous control problems in exchange for extra computation at training time but at no cost in computation at rollout time. Our approach integrates a mix between model-free and model-based reinforcement learning. Policy Prediction Network is the first to introduce implicit model-based learning to Policy Gradient algorithms for continuous action space and is made possible via the empirically justified clipping scheme. Our experiments are focused on the MuJoCo environments so that they can be compared with similar work done in this area.
\end{abstract}

\section{Introduction}\label{sec:intro}

Reinforcement learning algorithms can be
model-free or model-based. Model-free reinforcement learning attempts to find a policy through interacting with the environment and improving the policy based on previous states and rewards.
Model-based reinforcement learning attempts to learn the dynamics of the environment and uses the model to improve the policy
through various methods such as planning, exploration, and even training on generated data \cite{sutton1990integrated,ha2018recurrent}.
Though historically, model-based methods capable of predicting near perfect
observations
\cite{oh2015action,chiappa2017recurrent} usually have the benefit of reduced
sample complexity, they still struggle to perform as well as model-free methods
\cite{schulman2017ppo,mnih2015human,mnih2016asynchronous}. It is therefore appealing to explore
achieving the best of both worlds, as collecting the large amount of experience
required by model-free methods is oftentimes expensive or infeasible.

Model-based agents traditionally learn a model of the environment that predicts future
observations conditioned on previous actions and observations. This approach is sometimes referred to as
observation-prediction models \cite{oh2017vpn}. Recreating the original observation
is sometimes a questionable objective as the original observation can be dominated
by irrelevant information. For example, if the observation is an image and contains
a complex background, a large part of the model's capacity can be spent on modeling the background even though it may be irrelevant to information necessary for planning.

As a response to the issues faced by observation-prediction models, several
implicit model-based
methods \cite{oh2017vpn,Farquhar2017treeqn,Silver2016ThePE} were
introduced and learn an implicit transition module that predicts the value/reward of
future states without being subjected to observation reconstruction.
Value Prediction Networks (VPN)
\cite{oh2017vpn} and TreeQN \cite{Farquhar2017treeqn} operate by expanding a tree of predicted
reward and value estimates. However, this is feasible only because each branch is linked to a discrete action. ATreeC
\cite{Farquhar2017treeqn} introduces a policy gradient method but is still not applicable to continuous action spaces because their policy is a multinomial distribution parameterized by the Q-values associated with each branch.  Implicit
models have seen success in Q-learning approaches, but are not straightforward to
apply to policy gradient methods. Many real-world problems, such as robotics or autonomous vehicle applications,
lie in continuous action spaces and Q-learning approaches do not naturally extend to continuous action spaces like policy gradient methods.

Policy gradient methods are of primary interest in this paper because of their inherent flexibility in terms of their application to both discrete and continuous action spaces.
In particular, we will focus on a model-free policy gradient algorithm called Proximal Policy Optimization (PPO) \cite{schulman2017ppo}. PPO is of high interest because of its high performance on popular benchmarks and simplicity.

We propose Policy Prediction Networks (PPN), where the value, reward, policy, and abstract-state are predicted by leveraging a transition model.
PPN uses an implicit model-based approach at training time but a model-free approach at rollout time.
An implicit model-based approach at training time helps accelerate feature learning via predicting future policies, values, and rewards. All of which encourage the dynamics model to learn features that are well aligned with our objective of finding a policy that maximizes returns.
To the best of our knowledge, this is the first work on developing implicit model-based learning for policy gradient methods.

Our contribution is a training procedure that leverages model-based learning for policy
gradient algorithms to improve performance and does not trade off computational
costs at rollout time. This work introduces implicit transition models for Policy Gradient methods, depth-based objectives, auxiliary reward objectives, and an empirically justified clipping scheme.
Furthermore, our work lays down the foundation for future research on using implicit transition models to perform decision-time planning.
Empirical results demonstrate the advantage of PPN over
the model-free baseline (PPO),
which suggests that PPN finds a better state embedding and reduces sample complexity.

\section{Background and Related Work}\label{sec:related}
In this section, we will give a brief review of related work on model-based and model-free reinforcement learning. We also introduce
terminologies to differentiate between two general approaches in model-based reinforcement learning.

\textbf{Notations}: $s_t$ abstract state, $a_t$ action, $r_{t+1}$ reward from taking $a_t$ at $s_t$, $\gamma$ discount, $v_{\theta}(s_t)$ value of state $s_t$ with respect to parameters $\theta$, $A_t$ advantage, $H_t$ policy entropy, and the subscript $t$ denotes timesteps. 

\subsection{Policy Gradient Methods}
Policy Gradient Methods~\cite{sutton2000policy} are a type of reinforcement learning algorithm that directly optimizes policy parameters to maximize expected returns. 
Policy Gradient methods are more naturally applied to environments with continuous action spaces in comparison to Q-learning approaches. 
Generally, the policy gradient loss is of the shape:
\begin{align*}
\mathcal{L}^{\pi}_t=-\log\pi_{\theta}(a_t | s_t)A_t - H_t,
\end{align*}
where $A_t$ is an advantage estimate \cite{sutton2000policy,sutton2018reinforcement,schulman2015high}, $\pi_{\theta}$ is the policy with parameter $\theta$, $H_t$ is the policy entropy,
$s_t$ is the state at time $t$, and $a_t$ is the action taken in state $s_t$.

\subsection{Trust Region Policy Optimization}
Trust Region Policy Optimization (TRPO) attempts to generate monotonically improving policies following inspiration from conservative policy iteration \cite{kakade2002approximately}. 
However, TRPO contains a few theoretical relaxations that are required to make a practical algorithm.
TRPO is left with a hard KL divergence constraint to be less than or equal to $\delta$.
This hard constraint can be seen as a trust region on the mean KL divergence. 

Let $\theta'$ be the old parameters, $\theta$ be the new proposed parameters, $A^{GAE}_t$ be the generalized advantage estimate \cite{schulman2015high}.
Which is defined as
\begin{align}
    A^{GAE}_t = \delta_t + (\gamma \lambda) \delta_{t+1} + \dots + (\gamma \lambda)^{n - t + 1} \delta_{n-1}, \label{eq:gae}
\end{align}
where $\lambda$ is a hyperparameter controlling the bias-variance trade-off, and $\delta_t = r_{t+1} + \gamma v_{\theta'}(s_{t+1}) - v_{\theta'}(s_{t})$.
TRPO's optimization problem
is then formulated as:
\[ \pi_{\theta} = \max\limits_{\pi_{\theta}} L_{\pi_{\theta'}}(\pi_{\theta})
\;\; : \;\; \bar{D}_{KL}(\pi_{\theta'}, \pi_{\theta}) \leq \delta, \]
where 
$L_{\pi_{\theta'}}(\pi_{\theta}) =
\frac{\pi_{\theta}(a|s)}{\pi_{\theta'}(a|s)}A^{GAE}$ is the objective using importance sampling to estimate expected advantage under the new policy,
and $\bar{D}_{KL}(\pi_{\theta'}, \pi_{\theta})$ is the mean KL divergence between the new and old policy.
At this point TRPO offers theoretical inspiration but does not actually offer theoretical guarantees for monotonically improving policies.

\subsection{Proximal Policy Optimization}

Proximal Policy Optimization (PPO) \cite{schulman2017ppo} was introduced as offering similar benefits as TRPO \cite{schulman2015trust}, but via a simpler approach. 
PPO replaces a KL divergence constraint in TRPO via a clipped policy gradient loss:
\begin{equation} \label{eq:ppo}
\mathcal{L}^{\pi}_t = \max(-\text{ratio}_t \cdot A^{GAE}_t, -\text{ratio}_{t, \text{clip}} \cdot A^{GAE}_t),
\end{equation}
where
\begin{eqnarray}
    \text{ratio}_t & = & \frac{\pi_{\theta}(a=a_t | s=s_t)}{\pi_{\theta'}(a=a_t | s=s_t)}, \nonumber\\
    \text{ratio}_{t, \text{clip}} & = & \text{clip}\left(\frac{\pi_{\theta}(a=a_t | s=s_t)}{\pi_{\theta'}(a=a_t | s=s_t)}, 1-\epsilon,
    1+\epsilon\right). \label{eq:clip1}
\end{eqnarray}
Clipping no longer guarantees $\bar{D}_{KL}(\pi_{\theta'}, \pi_{\theta}) \leq \delta$. 
Instead, it serves to approximate it (see \cite{ilyas2018deep} for further details).

PPO2 \cite{baselines} is a GPU implementation from OpenAI that offers a key difference from PPO, namely, that the critic is also clipped. 
More specifically, the critic loss ($\mathcal{L}^{v}_t$) is now: 
\begin{align}
    \mathcal{L}^{v}_t = \max((v_{\theta}(s_t) - R_t )^2, (v_{t, \text{clip}} - R_t )^2)
\end{align}
Where $v_{t, \text{clip}} = \text{clip}(v_{\theta}(s_t) - v_{\theta'}(s_t), -\epsilon, \epsilon) + v_{\theta'}(s_t)$ is the clipped value estimate, $R_t = \gamma^{n} v_{\theta'}(s_{t+n}) + \sum_{i=1}^{n}\gamma^{i-1} r_{t+i}$ is the bootstrapped $n$-step return at time $t$, and $n$ is the number of steps in the bootstrapped estimate.
At this point theoretical guarantees in Conservative Policy Iteration have been dropped to make TRPO a practical algorithm, and the theoretical justifications in TRPO have again been weakened to make the more versatile and empirically superior PPO. 



\subsection{Model-based Reinforcement Learning}\label{ssec:model}
The essence of model-based reinforcement learning revolves around using a model or learning a model of the environment, and using this to improve a policy.
We will be focused on the challenging class of problems where the environment dynamics are unknown and must be learned. 
In this case, the problem can be broken down further into dynamics models that are learned implicitly and dynamics models that are learned explicitly.
It was not until recent years that implicitly learned model-based algorithms received attention \cite{oh2017vpn,Farquhar2017treeqn,Silver2016ThePE}.

\paragraph{Explicit Model-based Methods}
Cases of explicit model-based methods involve some form of directly predicting
future observations and including this in the loss function, as in:
\begin{align*}
& L^{model}_t = \frac{1}{2} \|\hat{x}_t - x_t\|^2,
\end{align*}
where $\hat{x}_t$ is the predicted observation at time $t$ and $x_t$ is the ground
truth observation.
Several variations exist that involve learning to predict in an abstract state space \cite{chiappa2017recurrent,ha2018recurrent,schmidhuber2015learning,pathak2017curiosity}, and predicting the grounded observation over multiple time steps \cite{oh2015action}. 
This has seen some success and can be useful for learning, planning, and exploration. 

These methods are particularly useful when observations contain entirely useful information.
However,
it can be misleading
in the class of problems where parts of the observation do not include
useful information.
For example, when generating an image frame, a
large part of the network's capacity could be dedicated to learning less useful
information like the background or objects that are not well aligned with the agent's interest~\cite{pathak2017curiosity}.

\paragraph{Implicit Model-based Methods}
Implicit model-based methods are interesting because they are not explicitly tied to reproducing original observations or an encoded observation. 
Rather, the dynamics model is indirectly learned by finding parameters that allow for an agent to perform optimally. 
This is done by learning to predict future characteristics such as the value or reward \cite{oh2017vpn,Farquhar2017treeqn}, but without having a constraint on predicting the ground truth observation. 
Unfortunately, a downside to implicit approaches is that it is difficult to know what is actually taking place during planning since it is hard to reconstruct the predicted observations.

VPN \cite{oh2017vpn} involve expanding a Q-tree, performing a linear backup along the maximal path, and selecting the maximal backed-up path. 
At training time, the loss is computed along the tree path followed by rollout actions.
The Predictron \cite{Silver2016ThePE} is similar to VPN except learning is done entirely in an abstract space, whereas VPN is grounded to transitions experienced by the rollout policy. 
Predictron also offers a meta-objective called consistency which lines up individual estimates with respect to backed-up estimates. 
We do not explore this in our work, but note that it could serve as an orthogonal improvement. 
TreeQN and ATreeC \cite{Farquhar2017treeqn} introduce a Q learning approach (TreeQN) and a policy gradient approach (ATreeC) that make use of a differentiable tree structure. 
ATreeC involves expanding a pseudo Q-tree.
This is pseudo because the nested value predictions are not directly constrained to represent the value. 
The backed-up pseudo Q-values are treated as logits and are used to parameterize and sample from a multinomial distribution. 
These samples are then used as the actions. 
VPN, ATreeC, and TreeQN are limited to only operating in discrete action spaces. 
Predictron was used in a continuous action space, the MuJoCo pool environment \cite{todorov2012mujoco}, but was done through discretizing the action space.

\section{Policy Prediction Network} \label{sec:ppn}
Policy Prediction Network uses a combination of model-free and model-based techniques. 
Actions are made with a model-free approach by the behavior policy at rollout time. 
However, learning is done with a model-based approach that follows the rollout trajectory. 
A latent space transition model is embedded into the architecture so that we are able to backpropagate from multiple simulation steps into the future back to a grounded observation. 
Backpropagation from predictions through the dynamics model, and back to a grounded observation enables the dynamics model to learn features which align with accurate reward predictions, accurate value predictions, and maximizing advantage. 
This is as opposed to maximizing observation reconstruction as is traditionally done in explicit model-based reinforcement learning. 

Our novel contribution is a training scheme that integrates model-free and model-based reinforcement learning to improve sample complexity and performance in exchange for extra computation at training time but at no extra cost in computation at rollout time. 
Additionally our work offers a foundation for decision-time planning for policy gradient methods and implicit transition models.
Our empirical results in Section \ref{sec:experiments} demonstrate the advantage of PPN over model-free baseline (PPO), which suggests that PPN finds a better state embedding and reduces sample complexity.

\subsection{Architecture} \label{ssec:arch}
\begin{wrapfigure}[16]{r}{0.5\textwidth}
    \centering
    \includegraphics[width=\linewidth]{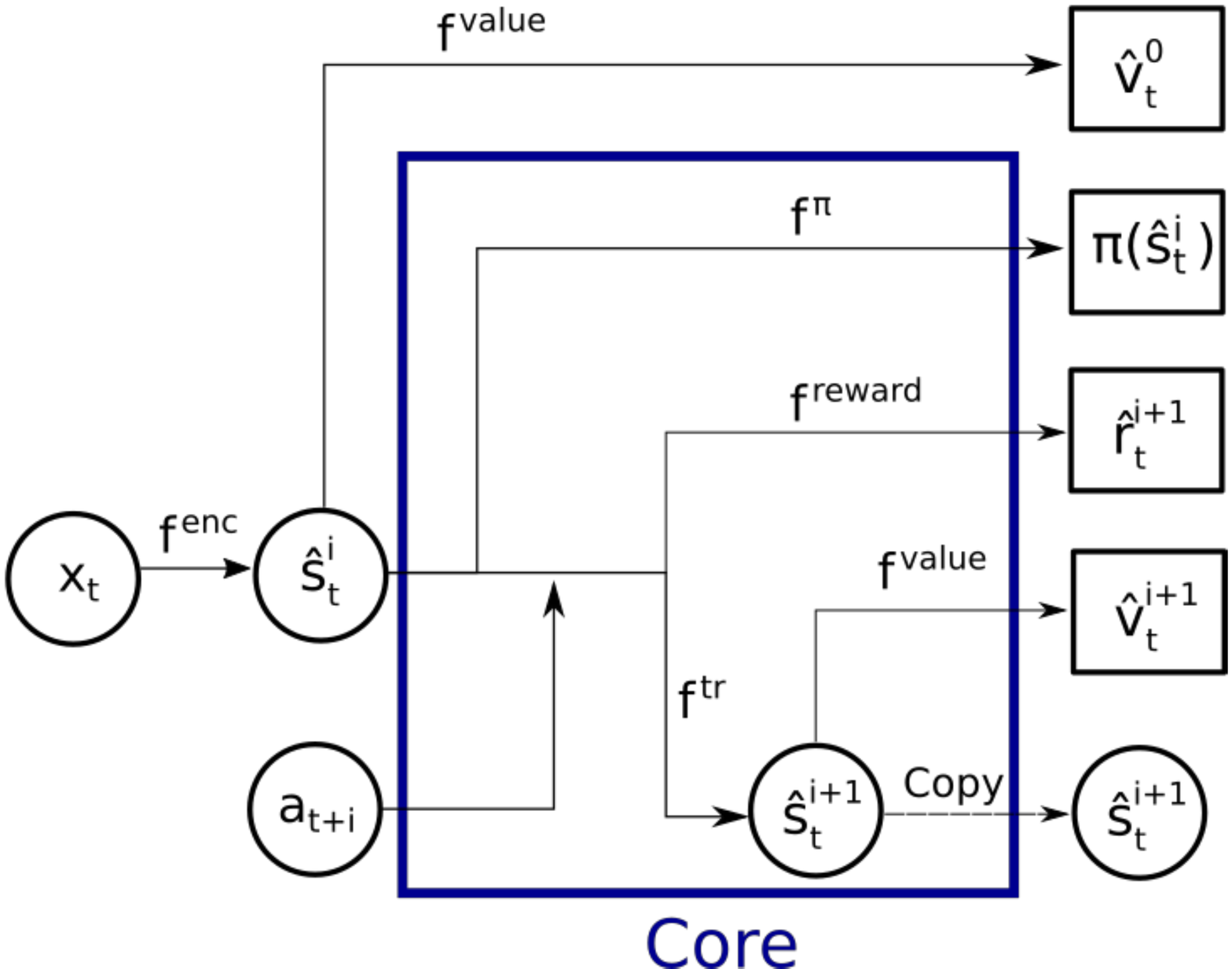}
    \caption{PPN learns to predict policies, rewards, abstract states, and the value of the abstract states.}
    \label{fig:arch}
\end{wrapfigure}
PPN is comprised of a few components. 
The components are parameterized by $\theta=\{\theta^{enc},\theta^{v}, \theta^{r},\theta^{tr}, \theta^{\pi}\}$ described below.
In the following, a hat over variables represents that it is an estimate as opposed to a grounded observation or reward. 
The superscript represents the forward step predictions.
The depth-rollout is expanded to a depth $d$. 
For example, $\hat{s}^i_t$ is the predicted state $i$ steps (where $0 \le i \le d$) forward in time from $t$.

\noindent
\textbf{Encoding} ($f^{enc}_{\theta}(x_t) = \hat{s}^0_{t, \theta}$) function embeds the observation ($x_t$) in an abstract state ($\hat{s}^0_{t, \theta}\in\mathbb{R}^y$).

\noindent
\textbf{Value} ($f^{v}_{\theta}(\hat{s}^i_{t, \theta}) =  \hat{v}^i_{t, \theta}$) function estimates the value ($\hat{v}^i_{t, \theta} \in\mathbb{R}$) of the abstract state.

\noindent
\textbf{Policy} ($\mathcal{N}(f^{\mu}_{\theta}(\hat{s}^i_{t, \theta}),
f^{\Sigma}(T))=\pi_{\theta}(\hat{s}^i_{t, \theta})$) function parameterizes a
distribution over actions to take given a state $\hat{s}^i_{t, \theta}$. The policy
module has two parts. The first ($f^{\mu}_{\theta}$) producing estimates of the
mean ($\mu \in\mathbb{R}^z$ where $z$ is
dimensionality of action space) and the second ($f^{\Sigma}$) producing estimates of a diagonal covariance matrix ($\Sigma \in\mathbb{R}^{z \times z}$) to parameterize a normal distribution for the policy ($\pi$). This is further described in Section~\ref{ssec:planning}.

\noindent
\textbf{Reward} ($f^{r}_{\theta}(\hat{s}^i_{t, \theta}, a_{t+i}) = \hat{r}^{i+1}_{t, \theta}$) function predicts the reward ($\hat{r}^{i+1}_{t, \theta} \in\mathbb{R}$) for executing the action $a_{t+i}$ at abstract state $\hat{s}^i_{t, \theta}$.

\noindent
 \textbf{Transition} ($f^{tr}_{\theta}(\hat{s}^i_{t, \theta}, a_{t+i}) = \hat{s}^{i+1}_{t, \theta}$) function transforms the abstract state given an action to the next abstract state ($\hat{s}^{i+1}_{t, \theta}\in\mathbb{R}^y$) by predicting $\Delta = \hat{s}^{i+1}_{t, \theta} - \hat{s}^i_{t, \theta}$.

We adopt a similar convention from VPN \cite{oh2017vpn} which defines a core module. Figure \ref{fig:arch} shows
the core module which performs a depth-1 rollout by composing the modules:
\begin{eqnarray*}
f^{enc}_{\theta}(x_t)& = & \hat{s}^0_{t, \theta}, \\
f^{v}_{\theta}(\hat{s}^0_{t, \theta}) & = & \hat{v}^0_{t, \theta}, \\
f^{core}_{\theta}(\hat{s}^i_{t, \theta}, a_{t+i}) & = & (\pi(\hat{s}^i_{t,
\theta}), \hat{r}^{i+1}_{t, \theta},\hat{v}^{i+1}_{t, \theta},\hat{s}^{i+1}_{t,
\theta}).
\end{eqnarray*}
There are 4 subtle but important differences between PPN depth rollouts and Value Prediction Network depth rollouts.
\begin{enumerate}
    \item PPN estimates the policy based on the abstract state ($\pi(\hat{s}^i_{t, \theta})$) at each step of the core module; while in VPN there is no need to predict a policy $\pi$ because it is a Q-learning method. However, this means it does not naturally apply to continuous action-spaces.
    \item PPN produces a value estimate ($\hat{v}^0_{t, \theta}$) at the base of the depth rollout and uses this as the critic. In VPN, this is not necessary because it is not an actor-critic method.
    \item The actions used in PPN come from samples generated by the behavior policy ($\pi_{\theta'}$), seen later in Equation (\ref{eq:sample}); while in VPN, the actions are chosen by exhaustively simulating all possible actions. Simulating all possible actions is only feasible in a discrete action space.
    \item PPN only uses the depth-based rollout at training time. VPN's behavior policy can use decision-time planning \cite{sutton2018reinforcement}. However, this is not straightforward to apply to continuous action spaces and we leave this for future work.
\end{enumerate}

If $d>1$, the PPN recursively calls the core function ($f^{core}_{\theta}$) to generate a trajectory of simulated rewards, policies, values, and abstract states conditioned on an initial abstract state ($\hat{s}^0_{t, \theta}$) and action trajectory ($a_{t}, \dotsc, a_{t+d-1}$). 
Each recursive call passes on the predicted abstract state ($\hat{s} = \hat{s}'$).

\subsection{Planning} \label{ssec:planning}
Here we introduce our approach to background planning \cite{sutton2018reinforcement} in continuous action spaces performed at training time. PPN has the ability to predict the future abstract states and based on these predicted future abstract states make additional predictions of future rewards, values, and policies. We use a basic planning method which simulates up to a certain depth $d$ collecting reward, value, and policy estimates along the way.

Background planning is done by following the actions performed by the behavior policy and recursively calling $f^{core}$ with the predicted abstract state. Action generation by the rollout policy ($\pi_{\theta'}$) is done by sampling from a normal distribution defined as follows:
\begin{align}
    a_t \sim N(\mu=f^{\mu}_{\theta'}(\hat{s}_{t, \theta'}^0), \;\Sigma=f^{\Sigma}(T)
	 | \hat{s}^0_{t, \theta'}, T).
    \label{eq:sample}
\end{align}
$f^{\Sigma}(T)$ is a function of the number of samples $T$ seen since the beginning of training, and does not depend on the model parameters. 
In our experiments the standard deviation used to parameterize a diagonal covariance matrix is exponentially decayed with respect to the number of samples seen, as is done in PPO \cite{schulman2017ppo}.



\subsection{Learning} \label{ssec:learning}
PPN is trained in a similar manner to typical Policy
Gradient algorithms. 
The novel differences we introduce are
depth-based 
losses, a latent transition model 
($f^{tr}$) embedded into the architecture, auxiliary reward objectives,
and a clipping scheme for depth-based losses.
Depth-based losses are necessary to train the implicit transition model. 
The implicit transition model and auxiliary reward
help with feature learning via background-planning.

PPN seeks to optimize auxiliary objectives and perform multiple updates on a batch.
Trust regions as seen in TRPO  can not be directly applied to both cases described above~\cite{schulman2017ppo} and thus we introduce a clipping approach.
Clipping all the network heads is crucial because the parameters of the reward network and value network all share parameters ($f^{\text{tr}}, f^{\text{enc}}$)
with the policy network.
For a visual reference of parameter sharing please see Figure~\ref{fig:arch}.
This means that if any of the networks are updated in an uncontrolled fashion, it can also cause dramatic changes to the policy. 

In addition, in Algorithm \ref{alg:learning}, we show that PPN performs a similar learning algorithm as was done in PPO. 
\begin{algorithm}[t]
  \caption{Policy Prediction Network(PPN), PPO style.}\label{alg:learning}
  \begin{algorithmic}
    \STATE Initialize parameters $\theta$
    \STATE $\theta' = \theta$
    \FOR{iteration=1, 2, $\dotsc$}
      \STATE Run policy $\pi_{\theta'}$ in environment for $n$ time steps
      \STATE Compute advantage estimates $A^{GAE}_1, \dotsc, A^{GAE}_n$
      \FOR{epoch$ =1$, $\dotsc$, $K$}
        \STATE Shuffle $n$ samples into mini-batches of size $M \leq n$
        \FOR{each mini-batch }
          \STATE $T$ is the set of samples selected for the mini-batch
          \STATE $\mathcal{L}_{mb} = \frac{1}{M}\sum_{t\in T} \mathcal{L}_t$
          \STATE Optimize $\mathcal{L}_{mb}$ w.r.t. $\theta$
        \ENDFOR
      \ENDFOR
      \STATE $\theta'=\theta$
    \ENDFOR
  \end{algorithmic}
\end{algorithm}

The major differences in training between PPN and PPO come from the loss formulation ($\mathcal{L}$).
The behavior policy
with parameters ($\theta'$) generates an $n$-step trajectory $(x_1,a_1,x_2,a_2,r_2,\dots,x_{n+1}, r_{n+1})$. 
The depth-$i$ predictions are grounded based on the generated $n$-step action trajectories. 
The loss at time $t$ accumulates error over the planned trajectory up to a depth $d$ and is defined as:
\begin{equation} \label{eq:loss}
\mathcal{L}_t = \mathcal{L}^{\pi}_t + \alpha_{v} \mathcal{L}^{v}_t + \alpha_r \mathcal{L}^{r}_t, \
\end{equation}
where minimizing $\mathcal{L}^{\pi}_t$ corresponds to maximizing expected advantage, $\mathcal{L}^{v}_t$ results in an accurate critic, and $\mathcal{L}^{r}_t$ leads to reward predictions that represent the environment's actual reward for a state action pair. 
$\alpha_{v}, \alpha_r$ are the penalty coefficients for the value loss and reward loss respectively.

Specifically, we define
\begin{equation}
\mathcal{L}^{\pi}_t = \frac{1}{d_{\pi}} \sum_{i=0}^{d_{\pi}-1} \max(-\text{ratio}^{i}_t A^{\text{GAE}}_{t+i}, -\text{ratio}^{i}_{t, \text{clip}} A^{\text{GAE}}_{t+i}) - \alpha_h H, \label{eq:pg_loss}
\end{equation}
where $\text{ratio}^i_t = \frac{\pi_{\theta}(a=a_{t+i} | s=\hat{s}^i_{t,\theta})}{\pi_{\theta^{'}}(a=a_{t+i} | s=\hat{s}^0_{t+i, \theta'})}$ is the importance sampling ratio between the new policy and the old policy at depth $i$, $\text{ratio}^i_{t, \text{clip}}$ is the clipped ratio used to ensure the new parameter's estimate to be near the old parameter's estimate.
We offer two possible formulations to clipping in Section~\ref{ssec:clipping}.
$A^{\text{GAE}}_{t}$ is the generalized advantage estimate defined in (\ref{eq:gae}),
and $\alpha_h$ is a hyperparameter for the entropy coefficient.

As for the critic objective, we have the critic loss
\begin{equation} \label{eq:vf_loss}
    \mathcal{L}^{v}_t = \frac{1}{d_v+1} \sum_{i=0}^{d_{v}} \frac{1}{2}\max((\hat{v}^{i}_{t, \theta} - R_{t+i})^2, (\hat{v}^{i}_{t, \text{clip}} - R_{t+i})^2), 
\end{equation} 
which encourages 
the current value estimate $\hat{v}^{i}_{t, \theta}$ to be close to the bootstrapped return $R_{t+i}$ at each depth $i$ without moving closer to the target than the clipped estimate ($\hat{v}^i_{t, \text{clip}}$).
The clipped estimate is guaranteed to be near the old parameter's estimate.
Notice that $\mathcal{L}^v_t$ is over an extra iteration of the summation. 
This is because value estimates are made at every state ($\hat{s}^0_{t, \theta}, \dotsc, \hat{s}^d_{t, \theta}$) in the forward plan. 

Similarly, the reward loss is
\begin{equation} \label{eq:r_loss}
\mathcal{L}^{r}_t = \frac{1}{d_r} \sum_{i=0}^{d_{r}-1} \frac{1}{2}\max((\hat{r}^{i}_{t, \theta} - r_{t+i})^2, (\hat{r}^{i}_{t, \text{clip}} - r_{t+i})^2), 
\end{equation} 
encourages the reward estimate $\hat{r}^{i}_{t, \theta}$ to be close to the reward $r_{t+i}$ at each depth $i$ without moving closer to the target than the clipped estimate ($\hat{r}^i_{t, \text{clip}}$).

The maximum in Equations (\ref{eq:pg_loss})-(\ref{eq:r_loss}) is taken between the unclipped surrogate objective and clipped surrogate objective.
In the case of the critic and reward losses
($\mathcal{L}^{v}_t$ and $\mathcal{L}^{r}_t$),
this means that updates only take place when the estimate from the new parameters
($\theta$) are farther from the target 
($R_{t+i}$ in Equation~\ref{eq:vf_loss} and $r_{t+i}$ in Equation \ref{eq:r_loss})
than the clipped estimate.
When the new parameter's estimate is closer, the max in Equation~\ref{eq:vf_loss} and~\ref{eq:r_loss} will select the clipped surrogate.
The gradient of the clipped surrogate with respect to parameters ($\theta$) will be zero,
and thus will not change any parameters.
This is desirable because it attempts to prevent destructive updates that push estimates made by $\theta$ far from estimates made by $\theta'$.

\begin{remark}
PPN can be reduced to PPO2 if ($\alpha_r=0$), $d_{\pi}=1$,
and $d_{v}=0$.
\end{remark}

It's possible to use different values of depth for the objectives but unless otherwise noted $d=d_{\pi}=d_v=d_r$.

\subsection{Clipping}\label{ssec:clipping}

We present two approaches to clipping called grounded
and ungrounded
clipping. 
In this case, grounded and ungrounded refer to whether we have access to the ground truth observation ($x_t$).
Grounded clipping offers a less strict clipping region,
while ungrounded clipping is more aligned with theoretical justifications found in Conservative Policy Iteration \cite{kakade2002approximately}, TRPO \cite{schulman2015trust}, and PPO \cite{schulman2017ppo}.
Our clipped objectives are advantageous for two reasons. First, they allow
for auxiliary reward and depth based updates. Second, they allow
us to share parameters between the transition, policy, value, reward, and embedding networks.
Both of these are essential
to learn the implicit transition model and are helpful with feature learning.

\subsubsection{Grounded Clipping}\label{sssec:grounded_clipping}

The clipping region is grounded with respect to both the action trajectory ($a_t, \dots, a_{t+d}$)
and the latent state ($\hat{s}^0_{t, \theta'}, \dots, \hat{s}^0_{t+d+1, \theta'}$). 
The three grounded clipped estimates are shown in Equations (\ref{ratio_clip_grounded}), (\ref{v_clip_grounded}), and (\ref{r_clip_grounded}):
\begin{eqnarray}
    \text{ratio}^i_{t, \text{clip}} &=& \text{clip}(\text{ratio}^i_t, 1-\epsilon, 1+\epsilon), \label{ratio_clip_grounded}\\
    \hat{v}^i_{t, \text{clip}} & =& \text{clip}(\hat{v}^i_{t, \theta} - v^0_{t+i,
	 \theta'}, -\epsilon, \epsilon) + \hat{v}^0_{t+i, \theta'}, \label{v_clip_grounded}\\
    \hat{r}^{i}_{t, \text{clip}} &=& \text{clip}(\hat{r}^i_{t, \theta} -
	 \hat{r}^0_{t+i, \theta'}, -\epsilon, \epsilon) +\hat{r}^0_{t+i, \theta'}. \label{r_clip_grounded}
\end{eqnarray}

The clipping region is based on the grounded estimates from the old parameters ($\theta'$) rather than predicted estimates from old parameters.

\subsubsection{Ungrounded Clipping}\label{sssec:ungrounded_clipping}

The clipping region in this case is grounded with respect to only the action trajectory,
but ungrounded with respect to the latent state.
The ungrounded clipping estimates are defined as:
\begin{eqnarray}
    \text{ratio}^i_{t, \text{clip}} & =& \text{clip}\left(\frac{\pi_{\theta}(a=a_{t+i} | s=\hat{s}^i_{t \theta})}{\pi_{\theta^{'}}(a=a_{t+i} | s=\hat{s}^i_{t, \theta'})}, 1-\epsilon, 1+\epsilon \right) \frac{\pi_{\theta^{'}}(a=a_{t+i} | s=\hat{s}^i_{t, \theta'})}{\pi_{\theta^{'}}(a=a_{t+i} | s=\hat{s}^0_{t+i \theta'})},\nonumber\\
    \hat{v}^i_{t, \text{clip}} & =& \text{clip}(\hat{v}^i_{t, \theta} - v^i_{t,
	 \theta'}, -\epsilon, \epsilon) + \hat{v}^i_{t, \theta'}, \label{v_clip_ungrounded}\\
    \hat{r}^{i}_{t, \text{clip}} & =& \text{clip}(\hat{r}^i_{t, \theta} -
	 \hat{r}^i_{t, \theta'}, -\epsilon, \epsilon) +\hat{r}^i_{t, \theta'}, \label{r_clip_ungrounded}
\end{eqnarray}
and $\text{ratio}^i_t =\frac{\pi_{\theta}(a=a_{t+i} | s=\hat{s}^i_{t, \theta})}{\pi_{\theta^{'}}(a=a_{t+i} | s=\hat{s}^0_{t+i, \theta'})}$.
Notice the change in how $\text{ratio}^{i}_{t, \text{clip}}$ is defined.
We first clip the ratio between new and old ungrounded policies to be no more or less $1 \pm \epsilon$ and then perform importance sampling to account for the advantage being calculated with respect to the rollout policy.

\begin{table}[t]
  \centering
  \begin{tabular}{ccc}
    \toprule
    & observation dimensions & action dimensions \\
    \midrule
    Hopper-v2                 & 11  & 3  \\
    Walker2d-v2               & 17  & 6  \\
    Swimmer-v2                & 8   & 2  \\
    HalfCheetah-v2            & 17  & 6  \\
    InvertedPendulum-v2       & 4   & 1  \\
    InvertedDoublePendulum-v2 & 11  & 1  \\
    Humanoid-v2               & 376 & 17 \\
    Ant-v2                    & 111 & 8  \\
    \bottomrule
  \end{tabular}
  \caption{Summary of the MuJoCo environments used.}
  \label{tab:envs}
\end{table}
\section{Experiments}\label{sec:experiments}
Our experiments seek to answer the following questions: 
(1) Is clipping necessary? If so, which type of clipping performs best(Section~\ref{exp:clip})?
(2) Does PPN outperform model-free baselines(Section~\ref{exp:baseline})?
(3) What effect does depth have on performance(Section~\ref{exp:depths})? 
(4) Is the implicit transition module actually predicting abstract states that are useful to the policy(Section~\ref{exp:ab})?

\subsection{Experimental Setup}
Our experiments investigate the comparison of PPN to PPO2 \cite{baselines} on the OpenAI Gym MuJoCo environments \cite{todorov2012mujoco,brockman2016openai}.
Preprocessing was done similarly to that of PPO \cite{schulman2017ppo}. 
Both PPO2 and PPN were implemented in Pytorch \cite{paszke2017automatic}.

The comparison against PPO2 is run over all 8 environments listed
in Table \ref{tab:envs}.
Due to returns being subject to high variance, we run tests over 15 seeds (which is 3-5 times more than the related works in \cite{schulman2017ppo,oh2017vpn}).
\footnote{Due to computational constraints InvertedDoublePendulum-v2 only uses 5 seeds}
Due to computational constraints in other experiments we run over 5 seeds on the Walker2d-v2 and Ant-v2 environments.

Our PPO2 implementation uses the same hyperparameters as the baselines implementation \cite{baselines}.
The largest difference in our PPO2 implementation is that we do not perform orthogonal initialization.
We did not include orthogonal initialization because it was not mentioned in the original PPO and we did not notice any clear performance benefits. 
For example, our implementation receives roughly double the returns on the HalfCheetah-v2 environment than the results reported by the baselines \cite{baselines} implementation of PPO2 using orthogonal initialization. 

Our PPN implementation uses similar hyperparameters:
2 fully connected layers for the embedding.
2 fully connected residual layers with unit length projections of the abstract-state ~\cite{Farquhar2017treeqn} for the transition module.
1 fully connected layer for the policy mean, value and reward.
All hidden layers have 128 hidden units and $\tanh$ activations.
In practice we use Huber losses instead of L2 losses, as was done in related implicit model based works~\cite{oh2017vpn}.

\subsection{Clipping} \label{exp:clip}
We first look into the effect grounded clipping the network heads has on returns gathered by PPN agents. 
Here we test to see if a strong policy can still be learned if the other network heads ($f^r, f^v$) are not clipped.
\begin{wrapfigure}[26]{r}{0.45\textwidth}
    \begin{subfigure}{\linewidth}
        \includegraphics[width=\linewidth]{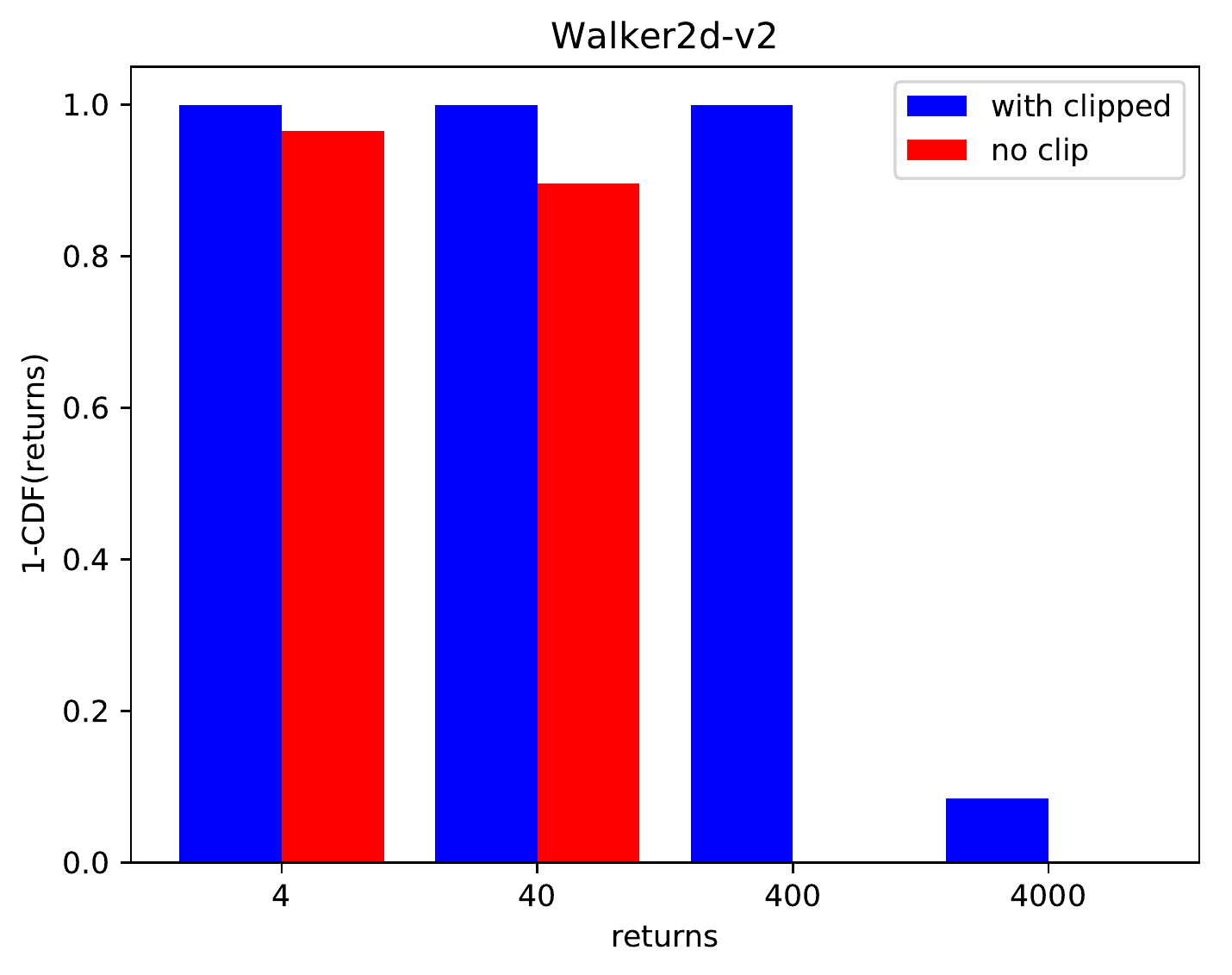}
        \caption{Walker2d-v2}
        \label{fig:clip_walker}
    \end{subfigure} \\
    \begin{subfigure}{\linewidth}
        \includegraphics[width=\linewidth]{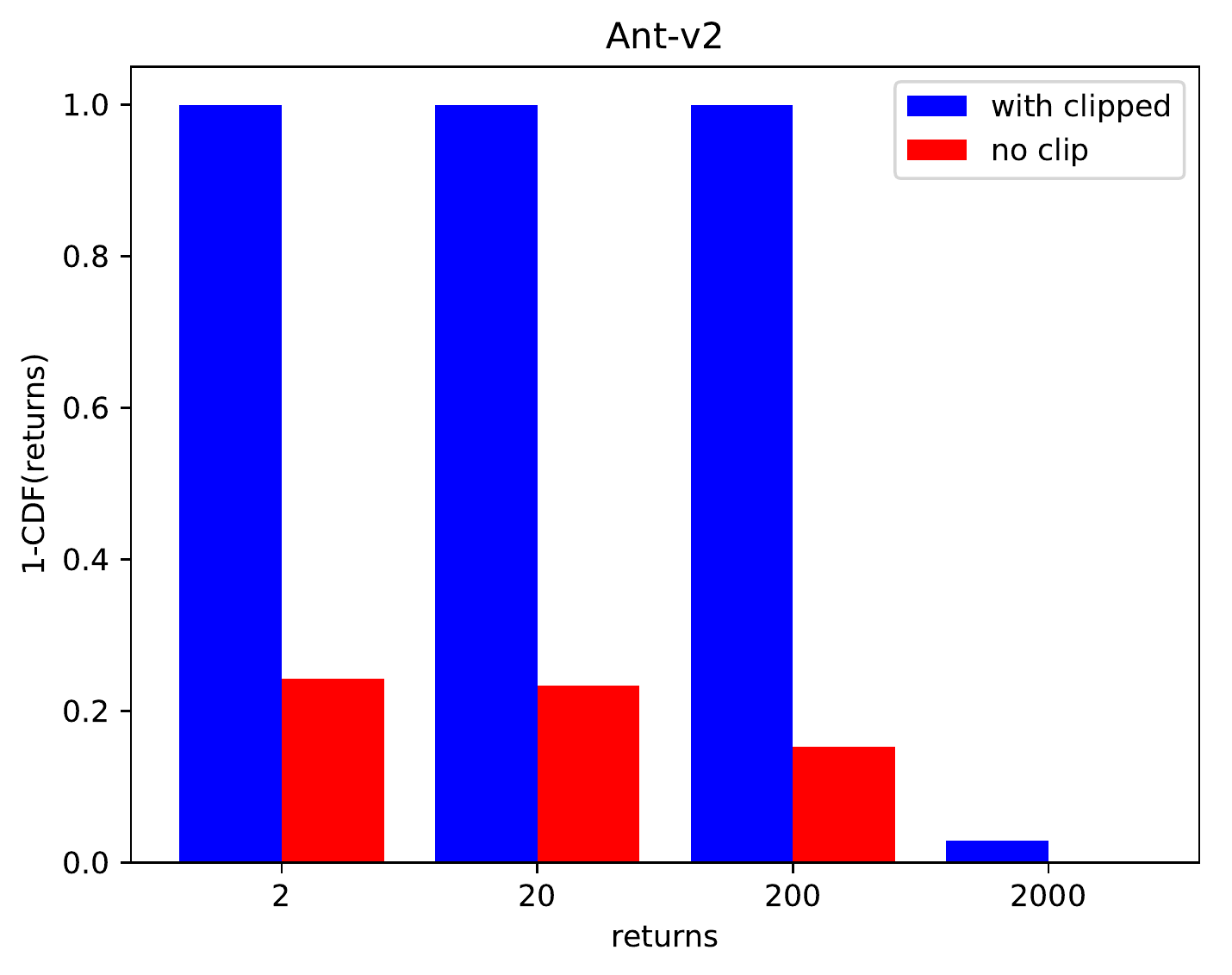}
        \caption{Ant-v2}
        \label{fig:clip_ant}
    \end{subfigure}
    \caption{Comparison of returns with and without (grounded) clipping of reward and critic.}  
	 \label{fig:clip}
\end{wrapfigure}

As similarly done by Ilyas \etal \cite{ilyas2018deep}, we fit a normal distribution to the returns achieved by the random seeds.
Then we compare points on the cumulative distribution functions (CDF) that correspond to returns of $2, 20, 200, 2000$ for Ant-v2 and $4, 40, 400, 4000$ for Walker2d-v2.

In Figure \ref{fig:clip_walker} and ~\ref{fig:clip_ant} we can see that clipping all the network heads turns out to be imperative to learn a useful policy. 
As stated in Section \ref{ssec:learning} clipping all the network heads is imperative because they all share parameters ($f^{\text{tr}}, f^{\text{enc}}$) with the policy.
Additionally, we look into which type of clipping performs best. 
For the most part Table \ref{tab:grounded_ungrounded} shows grounded clipping offers the most robust returns.
For all other PPN experiments we use the grounded clipping scheme.

\begin{figure}[t!]
    \begin{minipage}{\textwidth}
        \centering
        \begin{tabular}{cc}
            \includegraphics[height=3cm, width=.48\textwidth]{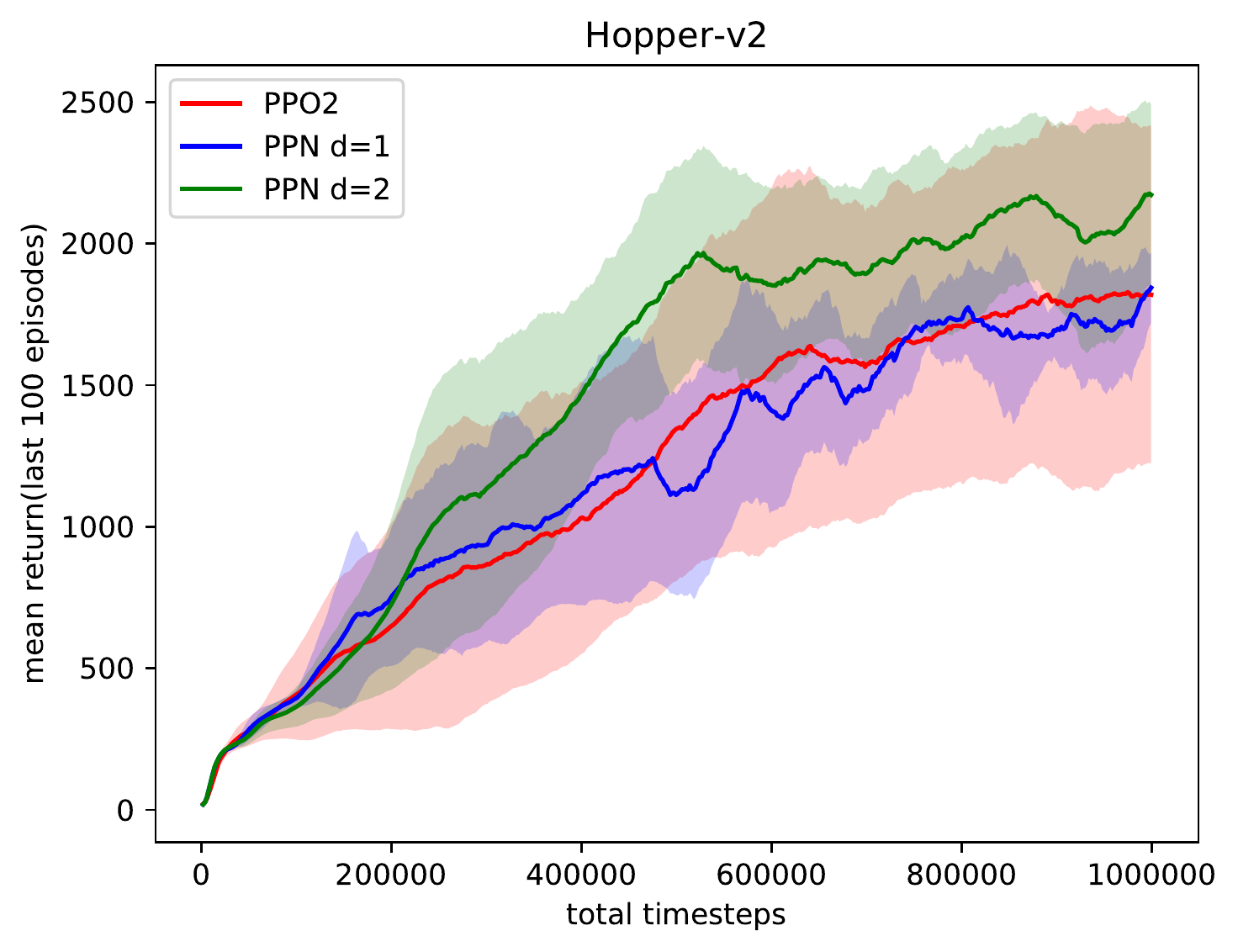}&
            \includegraphics[height=3cm, width=.48\textwidth]{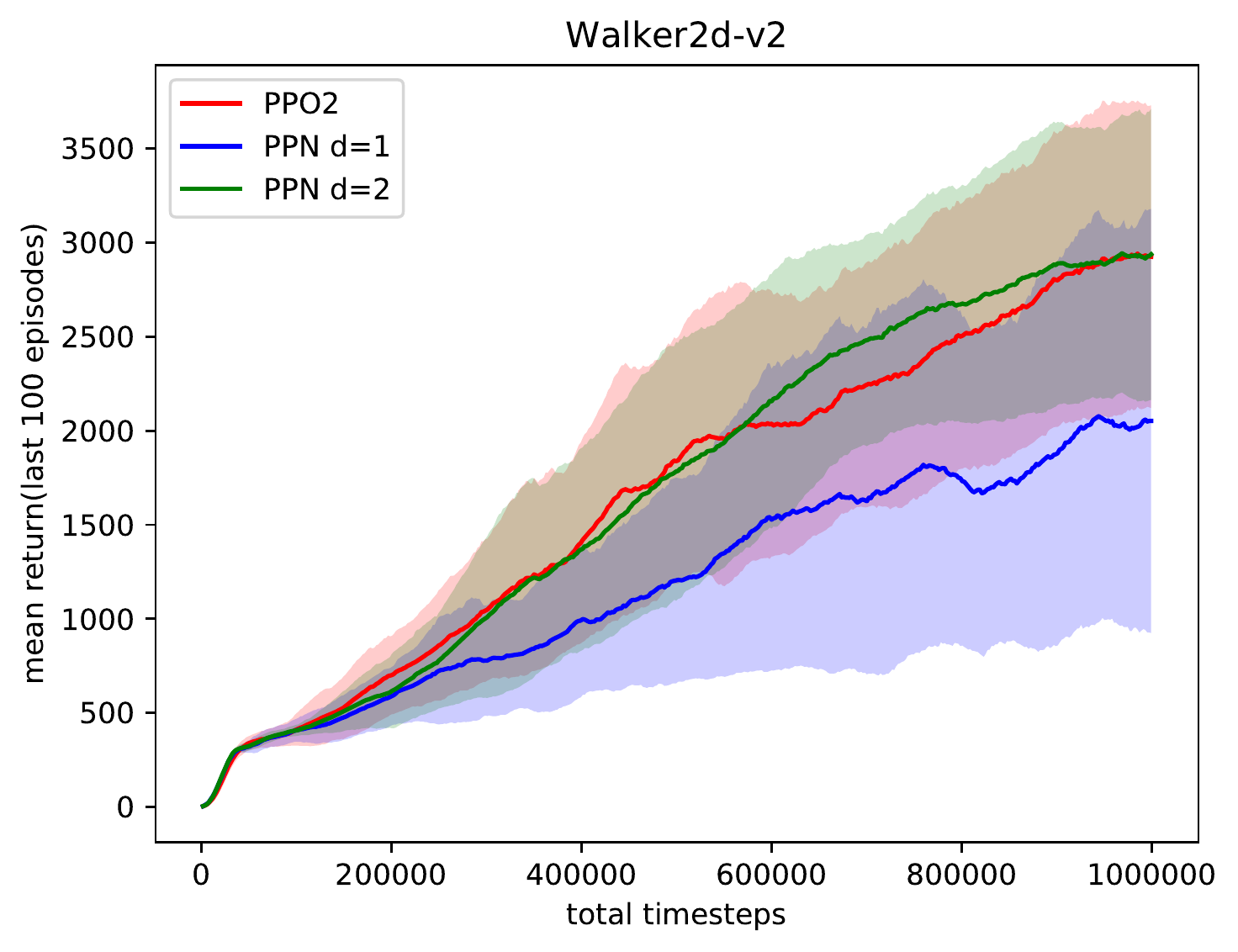} \\
            \includegraphics[height=3cm, width=.48\textwidth]{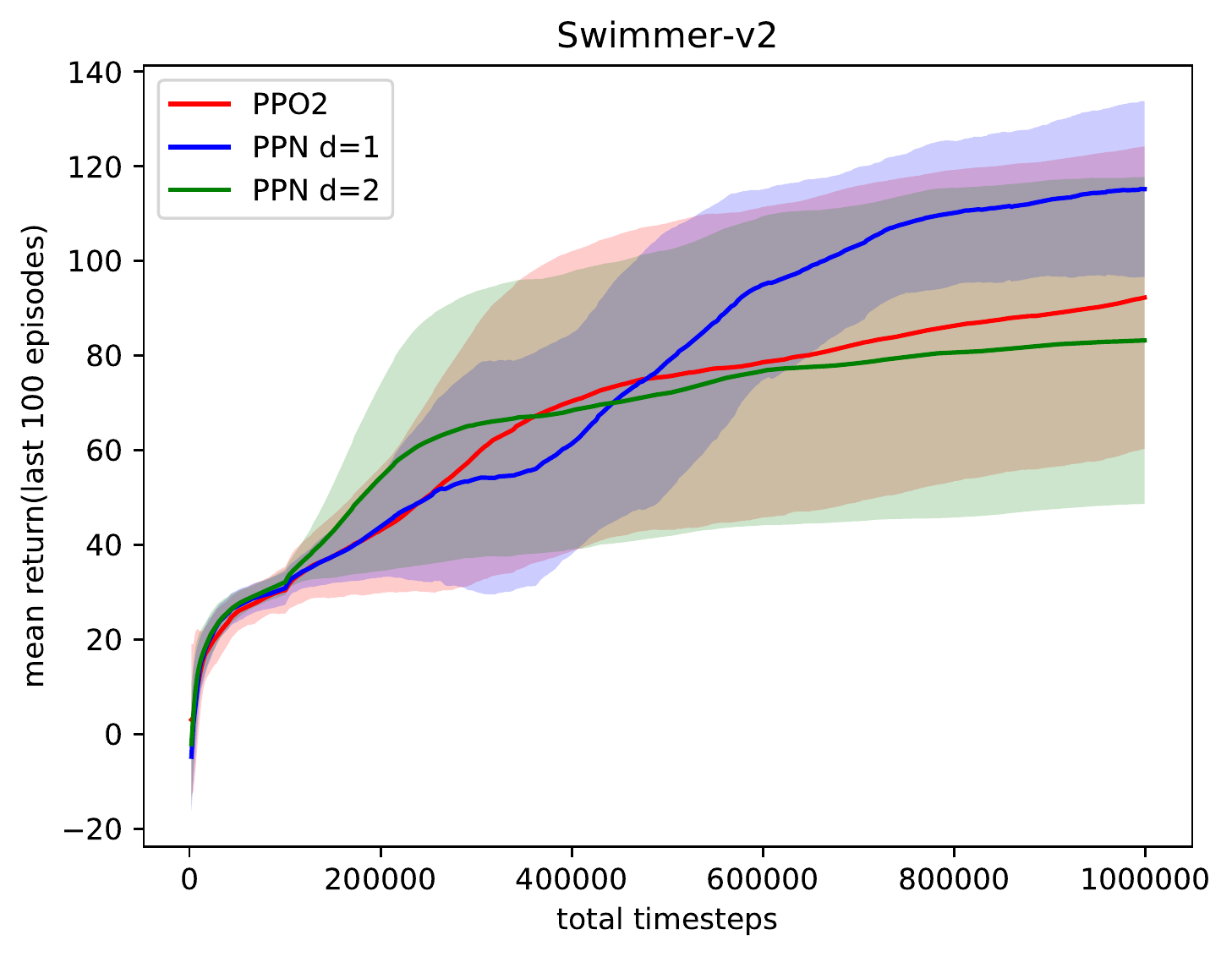} &
            \includegraphics[height=3cm, width=.48\textwidth]{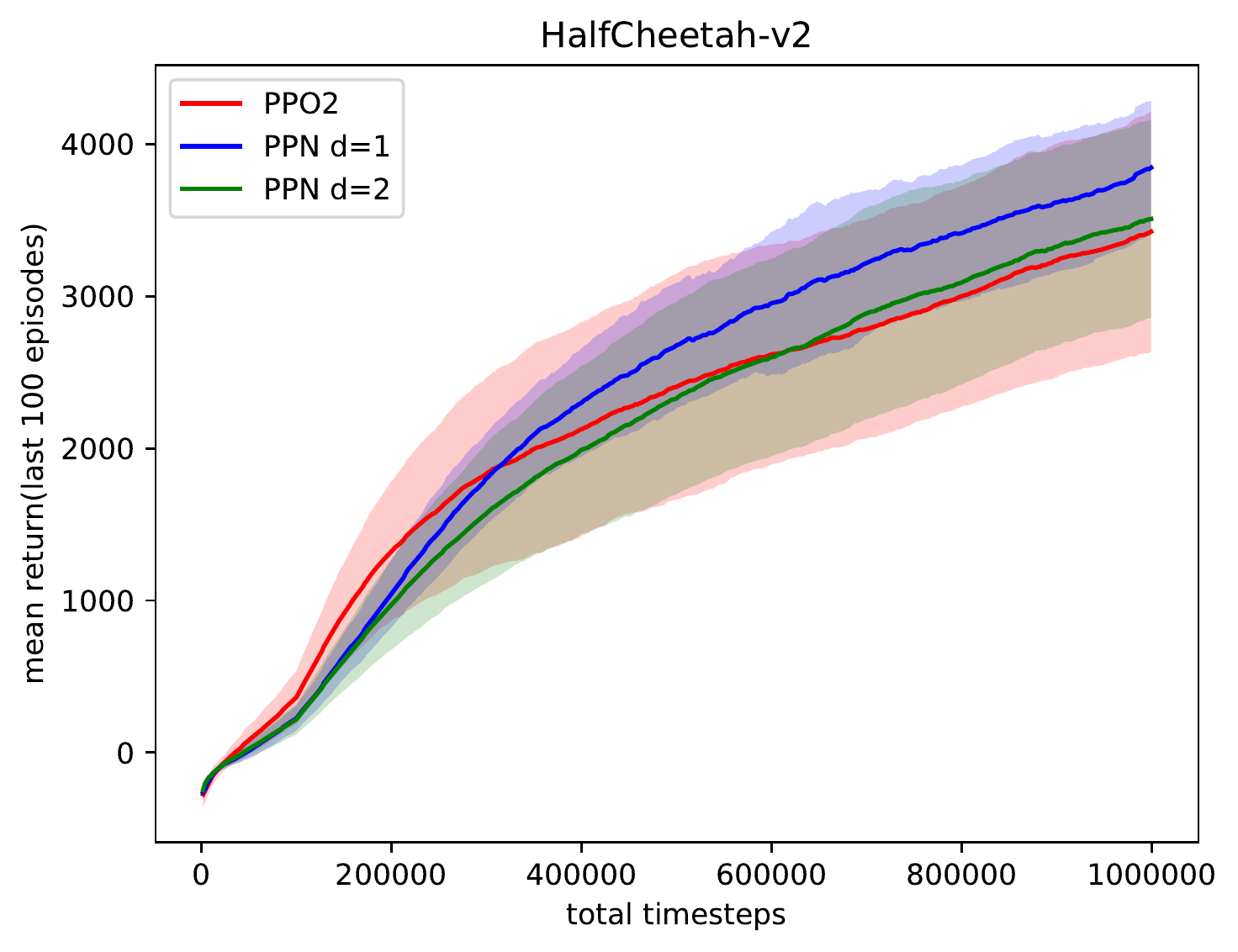} \\
            \includegraphics[height=3cm, width=.48\textwidth]{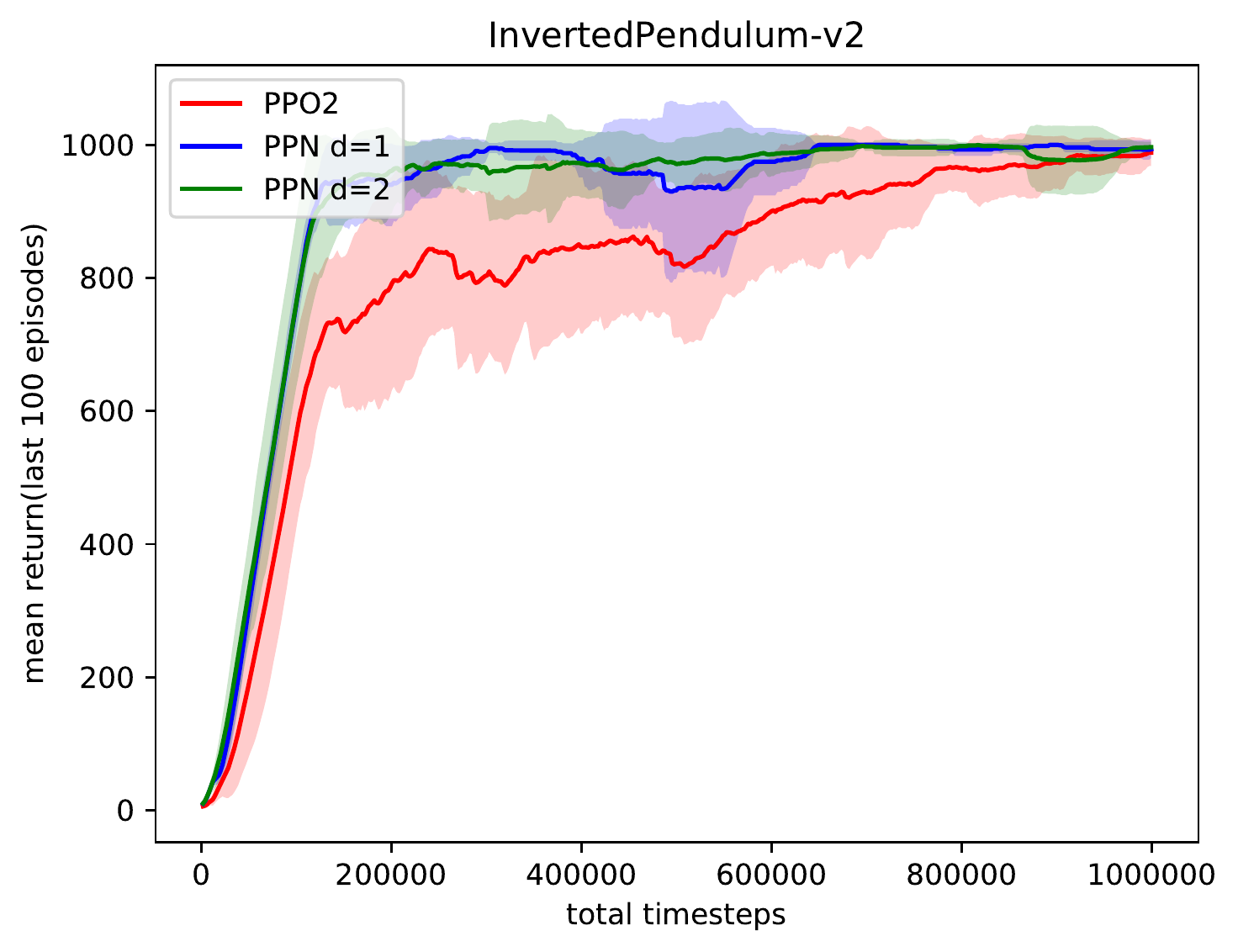}&
            \includegraphics[height=3cm, width=.48\textwidth]{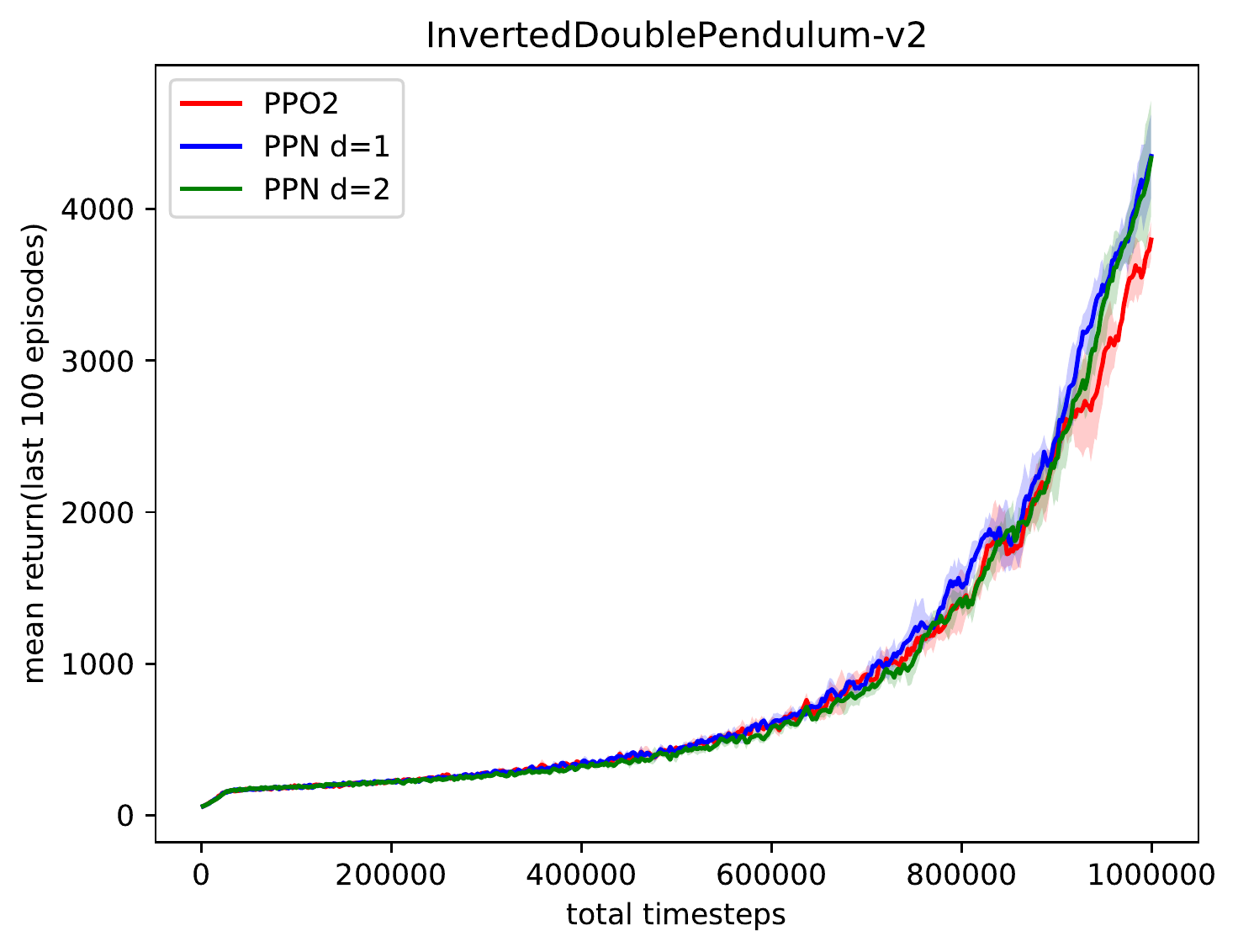} \\
            \includegraphics[height=3cm, width=.48\textwidth]{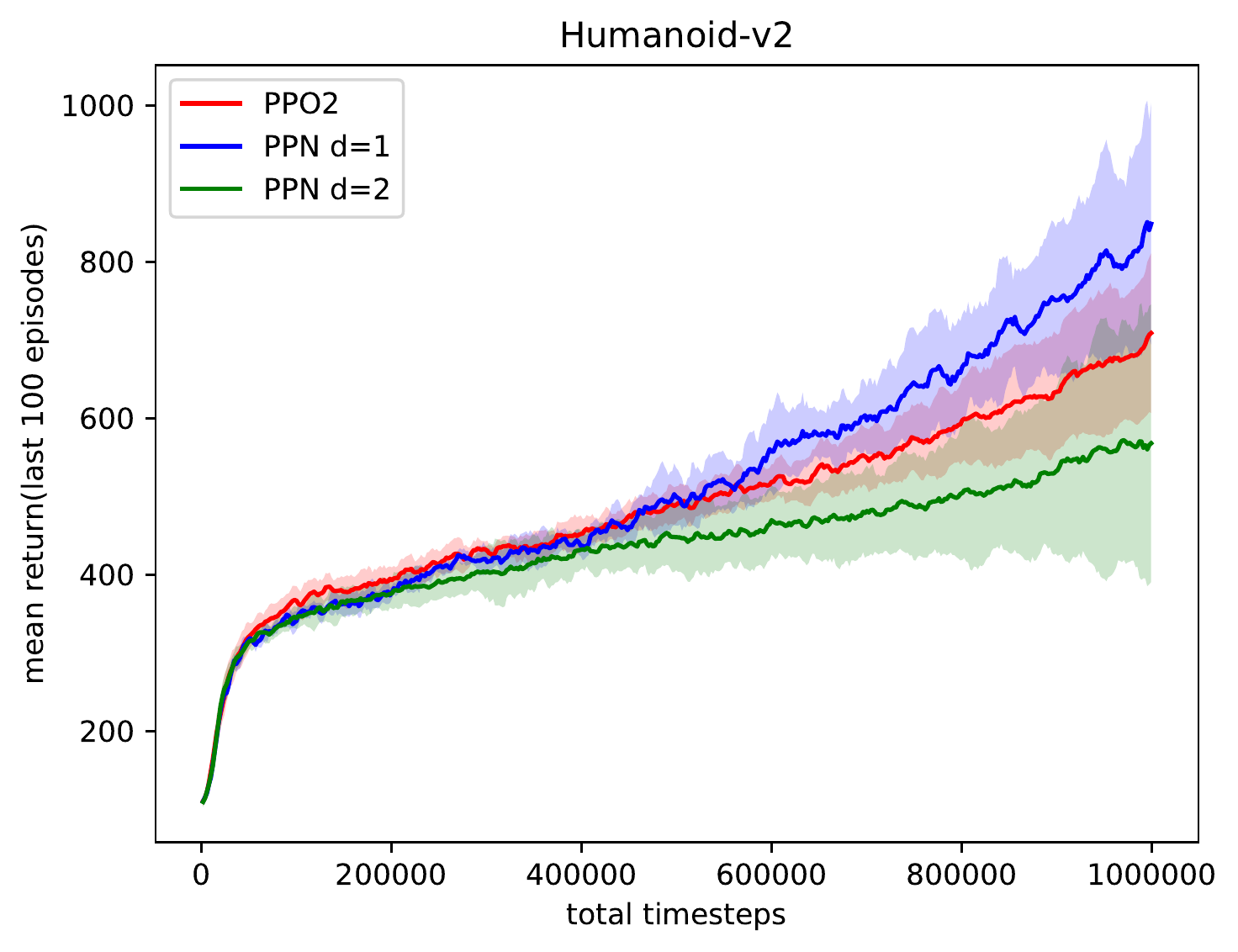}&
            \includegraphics[height=3cm, width=.48\textwidth]{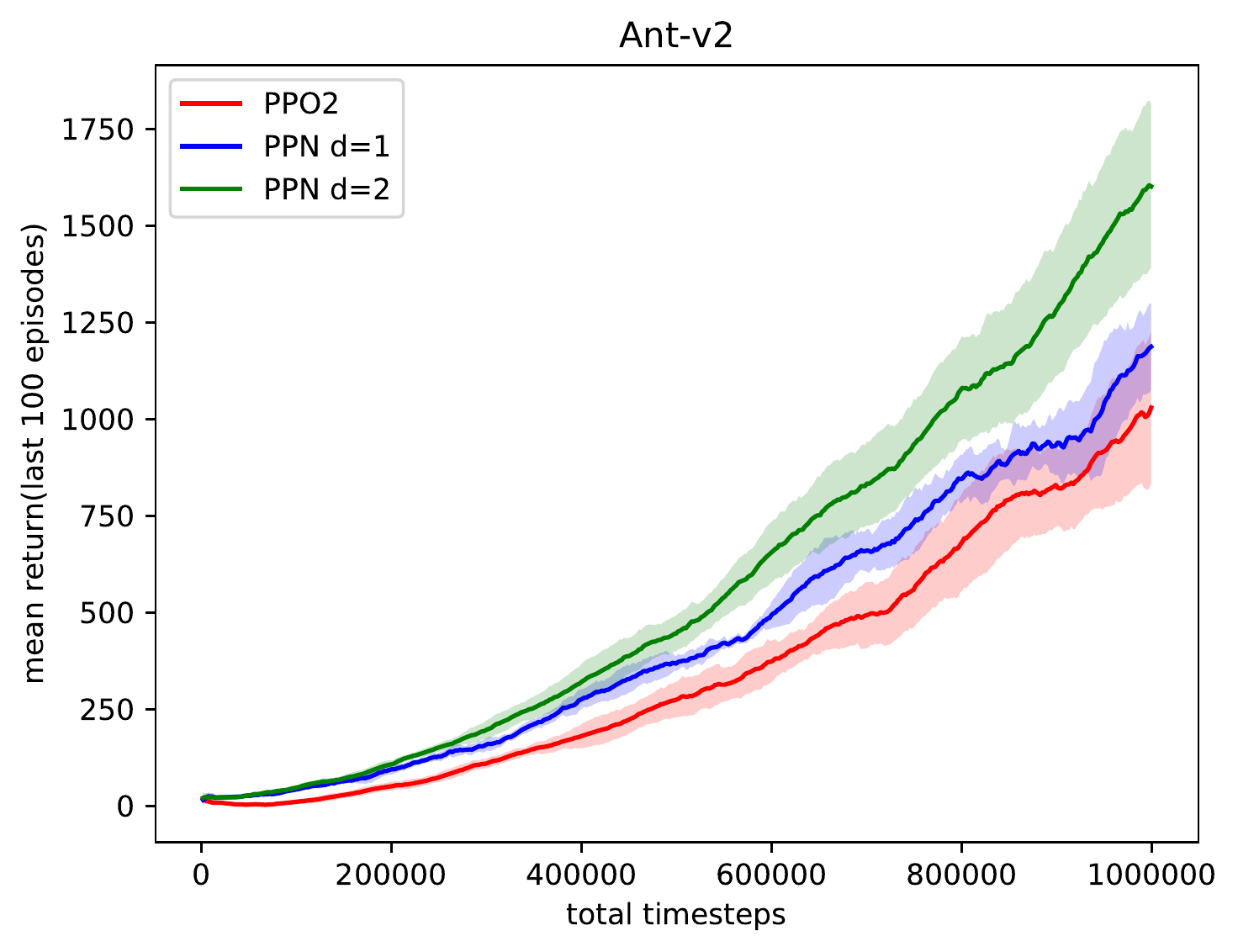}
        \end{tabular}
    \end{minipage}
    \caption{Results on 1 million step MuJoCo benchmark. Dark lines represent the mean return and the shaded region is a standard deviation above and below the mean.}
    \label{fig:mujoco}
\end{figure} 

\begin{table}[t!]
  \centering
  \begin{tabular}{cccc}
    \toprule
    & Grounded & Ungrounded \\ 
    \midrule
    Hopper-v2                 & 2172.28  & 1356.14 \\ 
    Walker2d-v2               & 2937.20  & 1717.23 \\ 
    Swimmer-v2                & 83.22    & 85.27   \\ 
    HalfCheetah-v2            & 3509.34  & 3485.59 \\ 
    InvertedPendulum-v2       & 996.44   & 998.47  \\ 
    InvertedDoublePendulum-v2 & 4336.93  & 4071.19 \\ 
    Humanoid-v2               & 574.15   & 676.31  \\ 
    Ant-v2                    & 1602.15  & 1566.06 \\ 
    \bottomrule
  \end{tabular}
  \caption{Returns using grounded and ungrounded clipping} 
  \label{tab:grounded_ungrounded}
\end{table}

\subsection{Baseline Comparison} \label{exp:baseline}
To test our model, we chose to benchmark against PPO2 and use the environments in Table~\ref{tab:envs}.
As is done in related works \cite{Farquhar2017treeqn}, we include depth $d=1$ and $d=2$ in our baseline comparison. 
However, we note that it is possible that larger depth values could be better on other environments.

As can be seen in Figure \ref{fig:mujoco}, we find that PPN finds a better if not comparable policy in all of the environments.
We notice that PPN tends to do well in complex environments such as Ant-v2. 
Indeed Humanoid-v2 is an exception to this observation. 
Perhaps this is because Humanoid-v2's number of observation dimensions(376) are far larger than the latent space(128). 
Additionally, we notice optimal depth is environment dependent and is further studied in Section~\ref{exp:depths}.

\subsection{Depth} \label{exp:depths}
In this section, we explore the effect depth has on returns gathered by PPN agents. 
Increasing depth forces the agent to learn abstract state which contains information relevant to longer-term environment dynamics.
\begin{figure}[t]
    \begin{subfigure}{0.48\textwidth}
        \includegraphics[width=\linewidth]{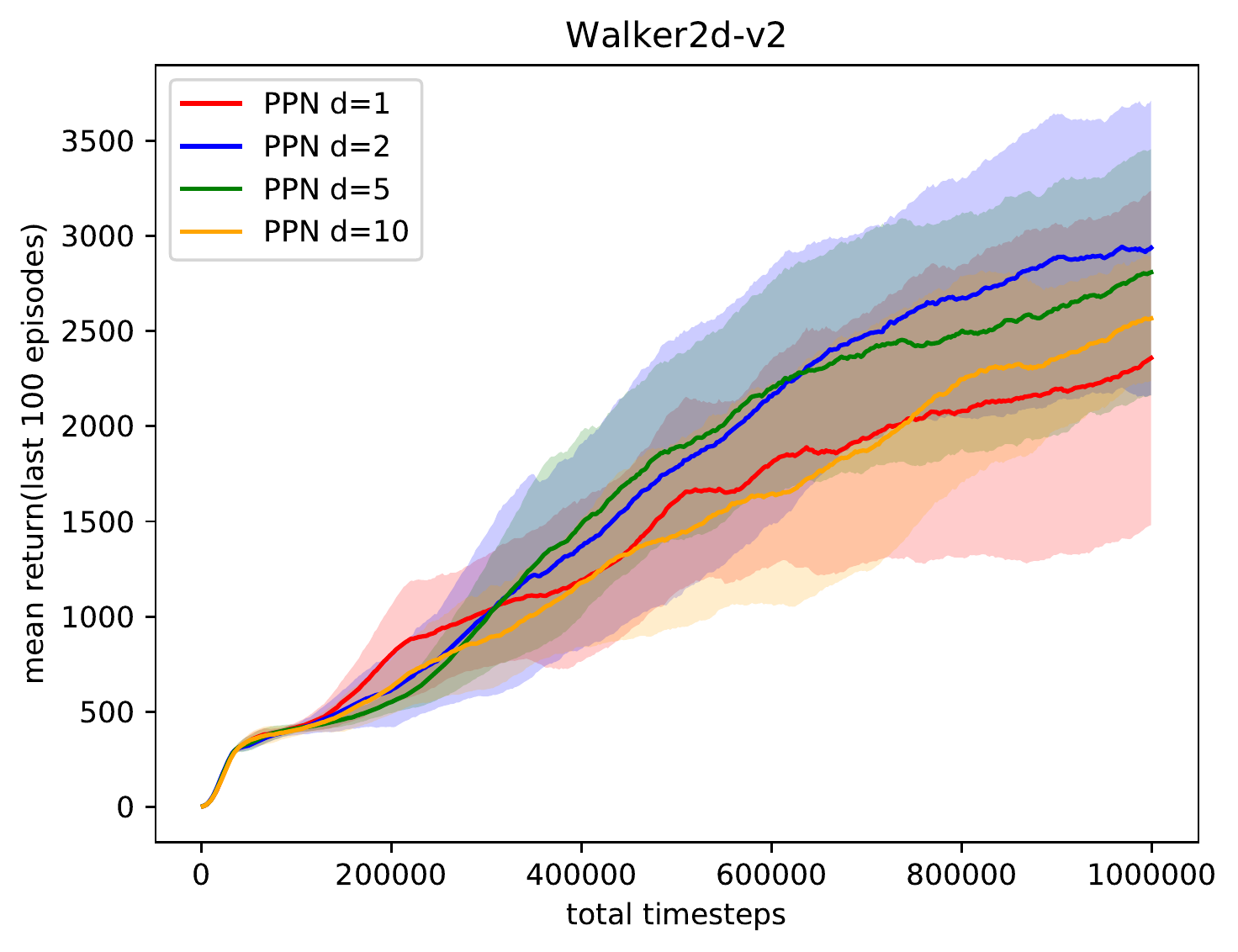}
        \caption{Walker2d-v2}
        \label{fig:depth_walker}
    \end{subfigure} 
    \begin{subfigure}{0.48\textwidth}
        \includegraphics[width=\linewidth]{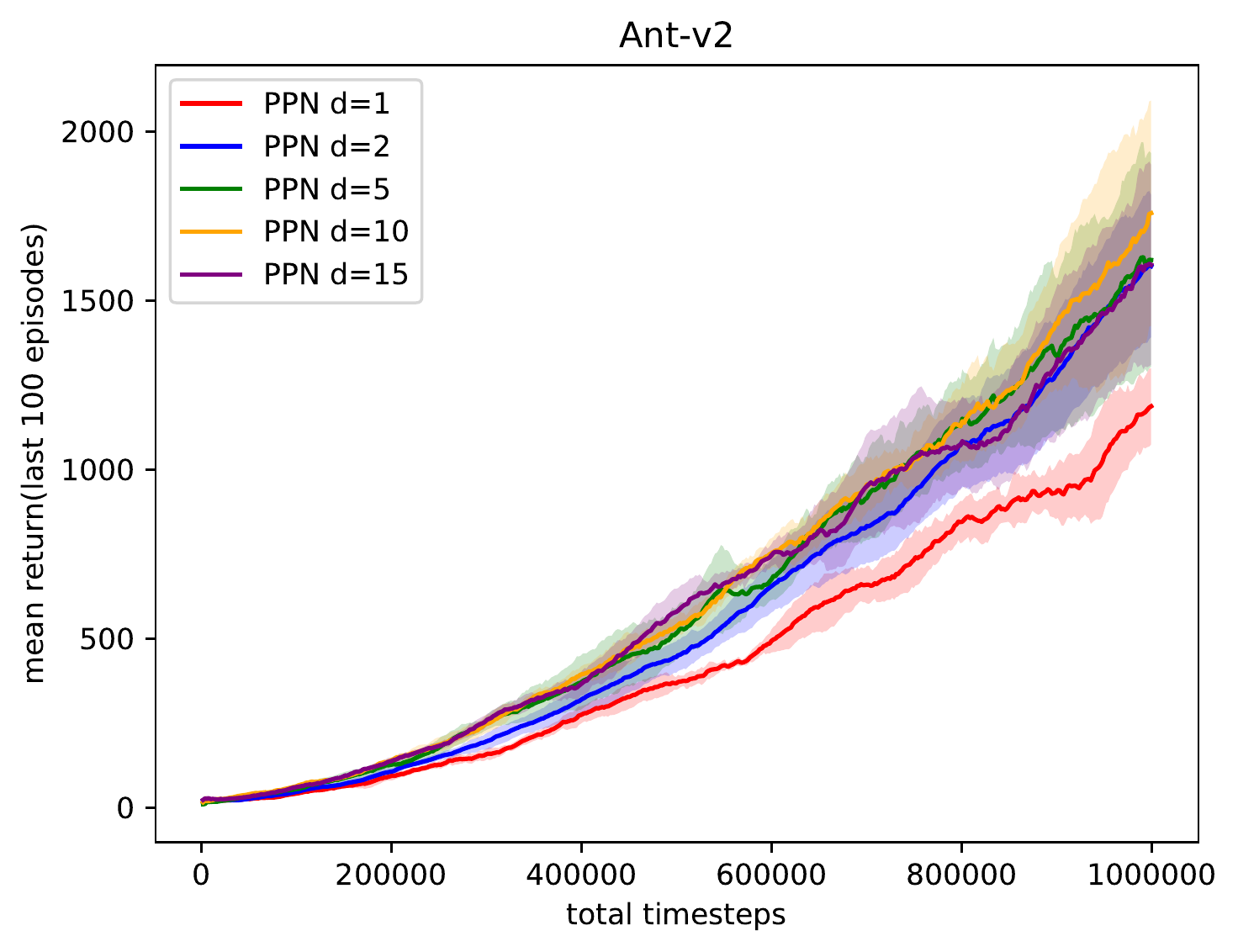}
        \caption{Ant-v2}
        \label{fig:depth_ant}
    \end{subfigure}
    \caption{Returns with respect to $d$ values of 1, 2, 5, and 10.}  
\label{fig:depth}
\end{figure}

As seen in Figure \ref{fig:depth} increasing depth ($d$) offers performance improvements but only up to a certain point.
As the depth grows, we become more reliant on having a good transition function and eventually leads to a worse policy. 

In Walker2d-v2 (Figure~\ref{fig:depth_walker}) we can clearly see a depth of 2 offers performance gains over a depth of 1.
However after this point returns decrease as we increase depth.
We suspect that optimal depth for Walker2d-v2 may be less than Ant-v2 because the implicit transition module is less accurate. 
A similar conclusion can be drawn from our observation in Section~\ref{exp:ab}.
Optimal depth is a recurring issue in implicit model-based approaches \cite{oh2017vpn,Farquhar2017treeqn}.

\subsection{Transition Ablation}\label{exp:ab}
\begin{figure}[t]
    \begin{subfigure}[b]{0.48\textwidth}
        \includegraphics[width=\linewidth]{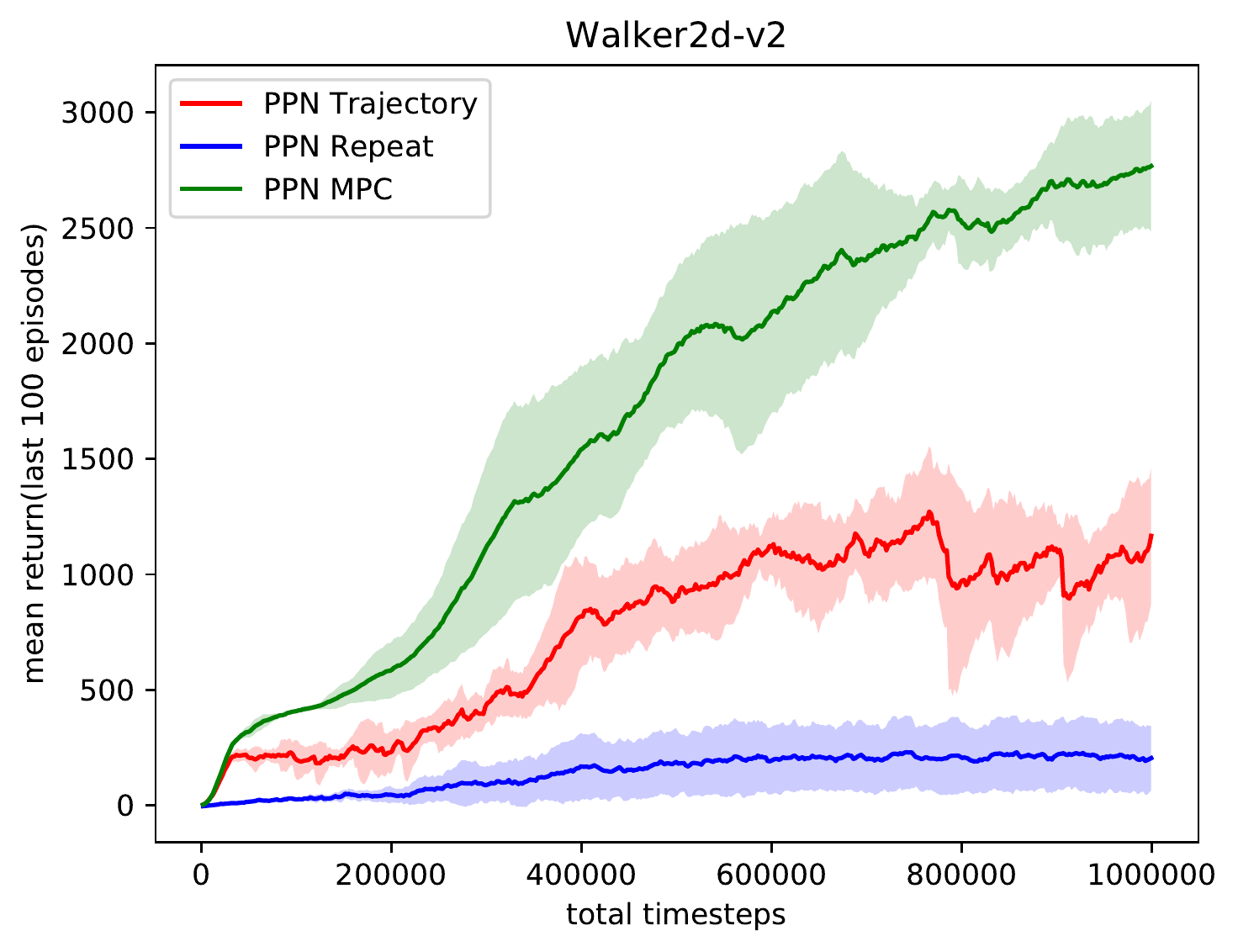}
        \caption{Walker2d-v2}
        \label{fig:walker_ab}
    \end{subfigure}
    \hspace*{\fill}
    \begin{subfigure}[b]{0.48\textwidth}
        \includegraphics[width=\linewidth]{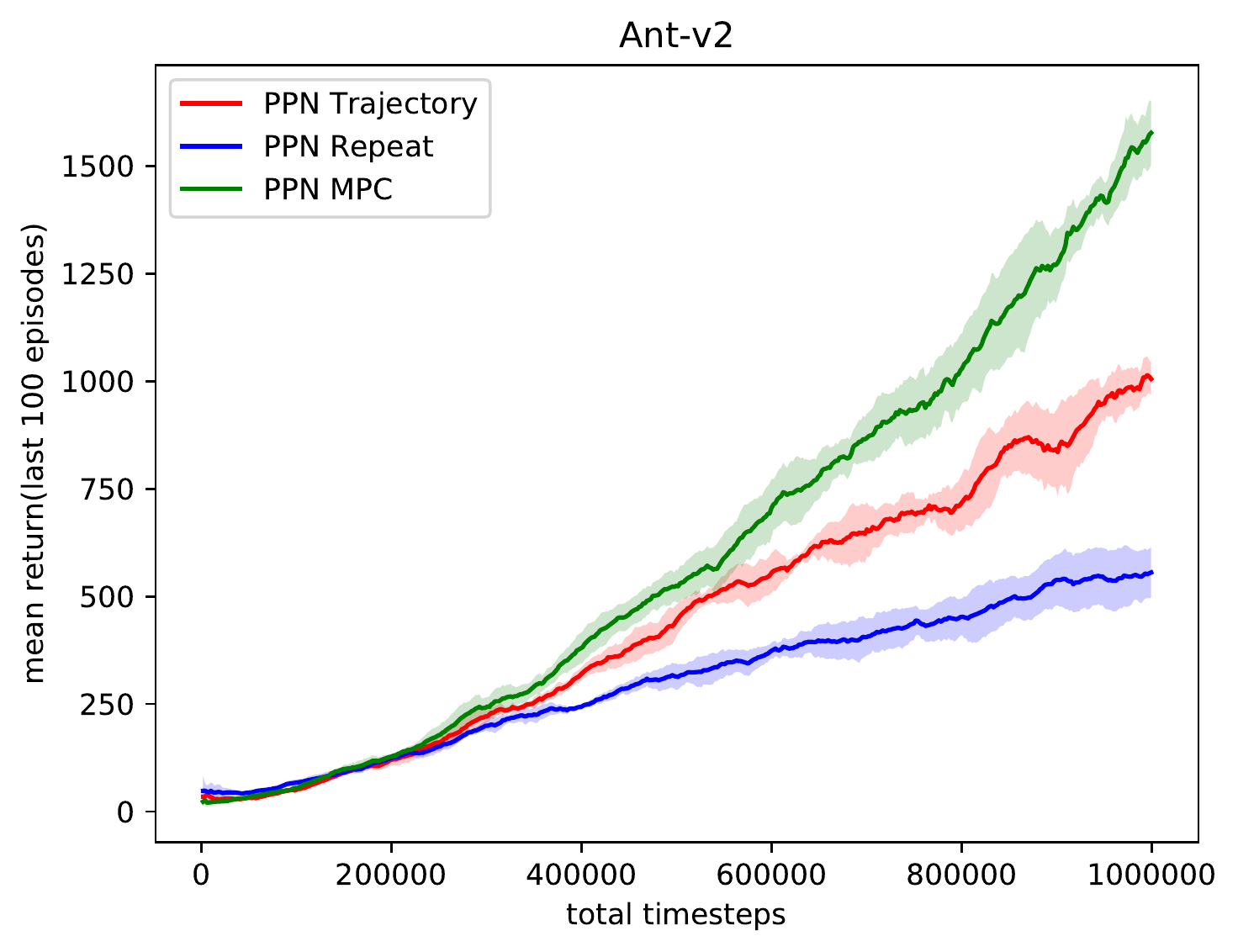}
        \caption{Ant-v2}
        \label{fig:ant_ab}
    \end{subfigure}
    \caption{Returns with respect to three difference action selection approaches.}  
\end{figure}
We are curious whether the implicit transition module is actually predicting abstract states that resemble reality closely enough to actually be useful to the policy. 
To test this we perform an ablation study of 3 different types of policy prediction networks. 
The first type Model Predictive Control (MPC) represents a perfect implicit transition module as the policy in this case has access to the ground truth observation. 
The standard MPC approach is where only the first action is followed and the rest are replanned. 
The second type, "trajectory", represents the strength of the transition module.  
In this case, every $d$ steps a new trajectory is generated by recursively calling $f^{\text{core}}$ with the predicted abstract states and the sampled actions the predicted policy.
The third type, "repeat", represents a meaningless transition module.
In this case, every $d$ steps a new action is generated by the policy and repeated for $d$ steps.
If the implicit transition module is bad we expect the returns from trajectory and repeat to be more or less the same. 
If the implicit transition module is good we expect returns somewhere in between the MPC and repeat curves. Note that all 3 of these approaches are trained in the same manner and have exactly the same parameters. 

In Figures \ref{fig:walker_ab} and \ref{fig:ant_ab} we see that the trajectory approach performs much better than repeat but not quite as well as MPC. 
This is interesting because the trajectory approach only has access to the grounded observation and must simulate $d$-steps into the future, where as in the MPC approach the action taken at time $t$ always has access to the observation from time $t$. 
These results show that the implicit transition module is indeed useful and could be used in future work for decision-time planning. 

\section{Conclusion}\label{sec:conclusion}
Introduced in this work is a learning scheme for Policy Gradient methods which integrates model-free and model-based learning that reduces sample complexity at no extra cost in computation at rollout time.
Additionally, PPN's implicit transition model acts as a first step towards decision-time planning with tree structured architectures in continuous action-spaces.
It is interesting to note that while we only explored continuous action spaces in this work it is also possible to extend this to discrete action spaces. 

For future work we would like to adapt PPNs to be less sensitive to planning depth and to leverage the transition model for decision-time planning. 
Decision-time planning is interesting but not straight forward to apply because it changes the behavior policy distribution in ways that are hard to measure.

\bibliographystyle{splncs04}
\bibliography{main}

\end{document}